\documentclass[letterpaper, 10 pt, conference]{ieeeconf}
\IEEEoverridecommandlockouts

\usepackage{pgfplots}
\pgfplotsset{compat=newest}
\usepackage{tikz}
\usetikzlibrary{calc}
\usetikzlibrary{shapes.geometric, arrows}

\usepackage{bm}

\usepackage{graphicx}	
\usepackage{caption}
\usepackage{subcaption}
\usepackage{amsmath}
\usepackage{amssymb}

\usepackage[noadjust]{cite}
\usepackage{balance}

\usepackage{renew}
\usepackage{mymath,diffge,gmech,lmech}

\title{Autonomous, Monocular, Vision-Based Snake Robot Navigation and
       Traversal of Cluttered Environments using Rectilinear Gait Motion}
\author{Alexander H. Chang$^{1}$, Shiyu Feng$^{1}$, Yipu Zhao$^{1}$, Justin S. Smith$^{1}$ and Patricio A. Vela$^{1}$%
\thanks{$1$: The authors are with the School of Electrical and Computer
Engineering, Georgia Institute of Technology, Atlanta, GA, USA.
Email: alexander.h.chang@gatech.edu, shiyufeng@gatech.edu, yipu.zhao@gatech.edu, jssmith@gatech.edu, pvela@gatech.edu}%
\thanks{This work was supported by NSF Awards \#1400256, \#1562911, and
\#1816138. Any opinions, findings, and conclusions or recommendations
expressed in this material are those of the author(s) and do not
necessarily reflect the views of the National Science Foundation.}
}

\newcommand{\boldpar}[2][0.05in]{%
  \vspace*{#1} \noindent \textbf{#2.}}

\begin{document}

\maketitle

\begin{abstract}
Rectilinear forms of snake-like robotic locomotion are anticipated to
be an advantage in obstacle-strewn scenarios characterizing urban
disaster zones, subterranean collapses, and other natural environments. 
The elongated, laterally-narrow footprint associated with these motion
strategies is well-suited to traversal of confined spaces and narrow
pathways. Navigation and path planning in the absence of global sensing,
however, remains a pivotal challenge to be addressed prior to practical
deployment of these robotic mechanisms. Several challenges related to
visual processing and localization need to be resolved to to enable
navigation. As a first pass in this direction, we equip a wireless,
monocular color camera to the head of a robotic snake. Visiual odometry
and mapping from ORB-SLAM permits self-localization in planar,
obstacle-strewn environments. Ground plane traversability segmentation in
conjunction with perception-space collision detection permits path
planning for navigation. A previously presented dynamical reduction of
rectilinear snake locomotion to a non-holonomic kinematic vehicle informs
both SLAM and planning. The simplified motion model is then applied to
track planned trajectories through an obstacle configuration. This
navigational framework enables a snake-like robotic platform to
autonomously navigate and traverse unknown scenarios with only monocular
vision.
\end{abstract}



\section{Introduction}
\seclabel{intro}
Navigation through cluttered environments, often describing natural
disaster aftermaths, subterranean collapses and similar senarios, is a
challenge that snake-like robotic platforms are well-positioned to
address. Mimicry of snake-like morphologies and adoption of serpentine
locomotion strategies potentially confer locomotive advantages similar to
those employed by biological counterparts traversing similar
environments. Given the current challenges associated to traversal of
arbitrary, unknown, rugged terrains, we focus the problem scope to
navigation through a unknown planar environments with obstacles.

A frequent characterization of these mission scenarios entails the
absence of global environmental knowledge. Success requires adequate
onboard sensing and navigation strategies in the presence of obstacles,
which includes self-localization, map construction and motion planning
\cite{SaEtAl_ICARCV[2016]}. Snake-like robotics research has focused on
the latter, path planning and following, with less effort on addressing
the former challenges.  Morphological features that advantage snake-like
platforms in obstacle-strewn environments also limit them with respect to
the supported onboard sensory equipment.
Head-mounted cameras obtain visual information which, in conjunction with
proprioception-based dead reckoning, informs localization
\cite{ZhEtAl_Sens[2018]}. Cameras on these robotic platforms have
predominantly been used to address higher task-level objectives or
monitoring needs rather than navigational autonomy
\cite{SaEtAl_ICARCV[2016], PaChKi_IROS[2000]}.  Other perceptual
modalities such as ultrasonic range sensors may be used to initiate
reactive obstacle avoidance \cite{TaKoTa_TCST[2015], PaChKi_IROS[2000],
WuEtAl_ROBIO[2012]}. Laser range finders have also been applied to
support SLAM for snake robots \cite{TiGoMa_ROBIO[2015],
TaKoTa_TCST[2015], WuEtAl_ROBIO[2012]}. 

Given a representation of the environment, path planning may be
accomplished. Approaches exploit simplified kinematic modes of travel for
hyper-redundant mechanisms. For active wheeled or tracked platforms,
follow-the-leader-like motion planning suffices \cite{PfoEtAl_RAS[2017],
FuWa_ICECC[2011], CaErEr_IROS[2007]}. Other planners focus on wheel-less
snake-like robots and exploit locomotive reduction of gaits to
differential-drive vehicle models for which control is well-understood;
in demonstration, however, they have presumed global knowledge of the
environment \cite{XiEtAl_ICRA[2015], HaEtAl_ICRA[2013], ChVe_ICRA[2017]}.
Integration of SLAM and path planning to accomplish navigation is
challenging for snake-like platforms due to weight as well as footprint
restrictions. Alternative vehicles operating terrestrially
\cite{ZhEtAl_IROS[2012], LaGeKi_ICRA[2011], GaEtAl_ICARSC[2016]},
aerially \cite{ChEtAl_ACC[2014], NuEtAl_JFR[2015]}, as well as
aquatically \cite{HiBr_ICARA[2015], HaRo_AUV[2014], Ki_UMich[2012]}, have
addressed similar navigational challenges, utilizing different sensing
strategies. However, mapping and self-localization, a key component for
autonomous navigation, remains largely unexplored on bio-inspired robotic
platforms.

\vspace*{0.25em}
\noindent {\bf Contribution.} 
We equip a snake-like robotic mechanism with a low-resolution, monocular camera.
This sensing modality, in conjunction with a reduced kinematic model of 
rectilinear motion, informs ORB-SLAM to accomplish self-localization in 
complex, obstacle-cluttered environments. 
Ground plane segmentation and ego-centric, perception space collision
checking facilitate trajectory planning under the assumption of a reduced
kinematic unicycle motion model. Lifting the reduced dynamics, and
corrective feedback modifications, to the actual locomotion gait leads to
execution of the planned trajectories. 
Relying on monocoluar vision, the presented framework enables a
snake-like robotic platform to both navigate and traverse unknown
environments.

\section{Rectilinear Gait Model}
\seclabel{gait}
\newcommand{\gshape}{g_{\text{shape}}}
\newcommand{\KECsym}{\mathbb{S}}

We briefly review a prior presentation of the traveling wave rectilinear gait and associated dynamics in \cite{ChSeVe_CDC[2016a], ChVe_ICRA[2017]}. 
The snake-like robot is represented as a mechanical system whose state decomposes
into a shape component, $r \in M$, and a group component, $g \in SE(2)$,
with body velocity, $\xi = \inverse{g} \dot g$. 
The reduced Lagrange-d'Alembert dynamical formulation is \cite{Ostrowski1996}:
\begin{equation} \eqlabel{mech_sys_eoms}
  \myvector{ \dot r \\ \dot g \\ \dot p}
  = 
  \myvector{u \\ 
            g \of{\body \Omega - \loc\Pconn(r, u)} \\ 
            \dual\ad_{\of{\body \Omega - \loc\Pconn(r, u)}} p
              + \body{\mathcal{F}}(r,p,u)}
\end{equation}
where $u$ is the shape space control signal, 
$\body \Omega$ is the vertical body velocity,
$p = \lock\li(r) \body \Omega$ is the vertical body momentum,
and $\loc \Pconn(r, \cdot)$ is the local principle connection defining
the horizontal and vertical split. The net external wrench acting on the body frame,
$\body{\mathcal{F}}$, models the influence of the external environment.

\subsection{Time-varying Gait Shape}
The time-varying gait shape for rectinilinear motion is formulated with
respect to an average body curve, parametrized by arclength, $s$, as well as
an accompanying average body frame rigidly attached to the average body at
midpoint, $s = 0$. Vertical lift of the gait shape from the ground is modeled 
in the $x-z$ plane, however, locomotion dynamics are modeled in the 
ground, $x-y$, plane. 

\begin{figure}[t!]
  \vspace{1.5mm}
  \centering
  \includegraphics[width=1.0\columnwidth,bb=0 0 1174 1014]{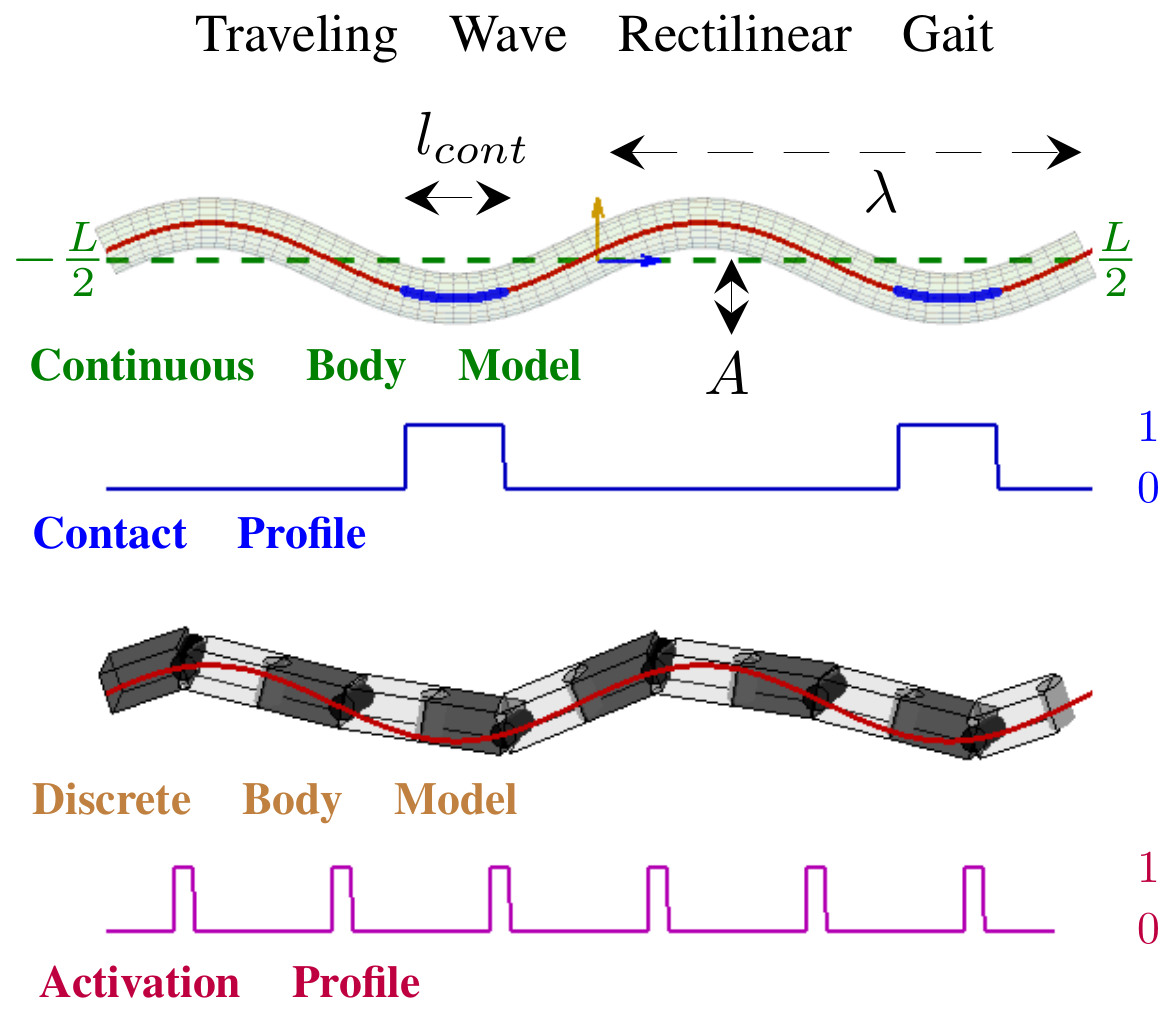}
  \vspace{-3mm}
  \caption{\textbf{Top:} Side view of the continuous body model
  annotated with the rectilinear traveling wave parameters. The green
  dashed line is the average body curve. Blue segments are overlaid onto 
  the body curve (red), identifying segments in ground contact.
  Pulses in the contact profile (blue) correspond to
  body segments in contact with the locomotion surface. \textbf{Bottom:}
  A discrete-link structure approximates the continuous body curve.
  Pulses in the activation profile (purple) depict regions of the body
  capable of effecting curvature and, thus, forward propulsion. All other
  regions may only induce drag.
  \label{rectilinear_body_curve_model}}
  \vspace{-6mm}
\end{figure}

A sinusoidal wave traveling rostrally along the body in the $x-z$ body
plane models the rectilinear gait, with time
varying shape equations,
\begin{equation} \eqlabel{travel_wave_body_shape}
\begin{split}
  d(s, t) & = \myvector{x(s) \\ y(s)} 
    = \frac{1}{\kappa} \myvector{\sin(\kappa s) \\ \cos(\kappa s) -1},  \\
  z(s,t) & = A \sin\of{2 \pi \of{f t + \frac{s}{\lambda}}},
\end{split}
\end{equation}
where $s \in [-L/2, L/2]$ is the body curve parameter 
(see Figure \ref{rectilinear_body_curve_model})
and $\kappa$ captures planar curvature of the average body (dashed green) in 
the $x$-$y$ modeling plane, assumed constant along the length of the body.
The variables $A$, $\lambda$ and $f$ parametrize the traveling wave
amplitude, wavelength and frequency, respectively while $t$ denotes time.  

Figure \ref{rectilinear_body_curve_model} (top) illustrates the continuous
body model, in the $x$-$z$ plane. A contact profile (beneath), parametrized by $s$, defines time varying locations on the body in contact with ground.
The $x$- and $y$- (into the page) 
coordinate vectors comprise the rigid  body frame with respect to which body frame dynamics are modeled, in the locomotion plane.

The rigid multi-link structure of the robotic snake (lower snake
image in Figure \ref{rectilinear_body_curve_model}),
in contrast to a continous body model, may only
effect curvature about joint locations along the body and not when the flat
face of the discrete link is flush to the ground. To accomodate this 
distinction an activation profile $1_{\rm act}(s) \in \{0,1\}$ is introduced, 
illustrated in Figure \ref{rectilinear_body_curve_model} (bottom). 
Pulses of the profile coincide with joint locations along the body which can
curve or roll and produce propulsion through a rolling friction model 
with the environment. All other regions of the body
coincide with link interiors, incapable of effecting curvature; they 
instead produce drag.

\subsection{Dynamical Model Reduction}
The traveling wave rectilinear gait has sufficient structure that the
equations of motion simplify. We advance to the modeling results most relevant to this work and refer the reader to \cite{ChSeVe_CDC[2016a], ChVe_ICRA[2017]} for specifics. 

A rolling viscous friction model is employed to model environmental caudal-rostral forcing produced at points of body-ground contact. Given the friction coefficients $\mu_b$, $\mu_f$, and $\mu_t$, individual link masses, and the contact and activation profiles, numerical
integration of the system dynamics permits recovery of the body velocity of
the robot as a function of time for a particular set of gait parameters.

\begin{figure}[t!]
  \vspace{1mm}
  \centering
  \resizebox{1.0\columnwidth}{!}%
    {\includegraphics[width=1.0\columnwidth,bb=0 0 1229 1001]{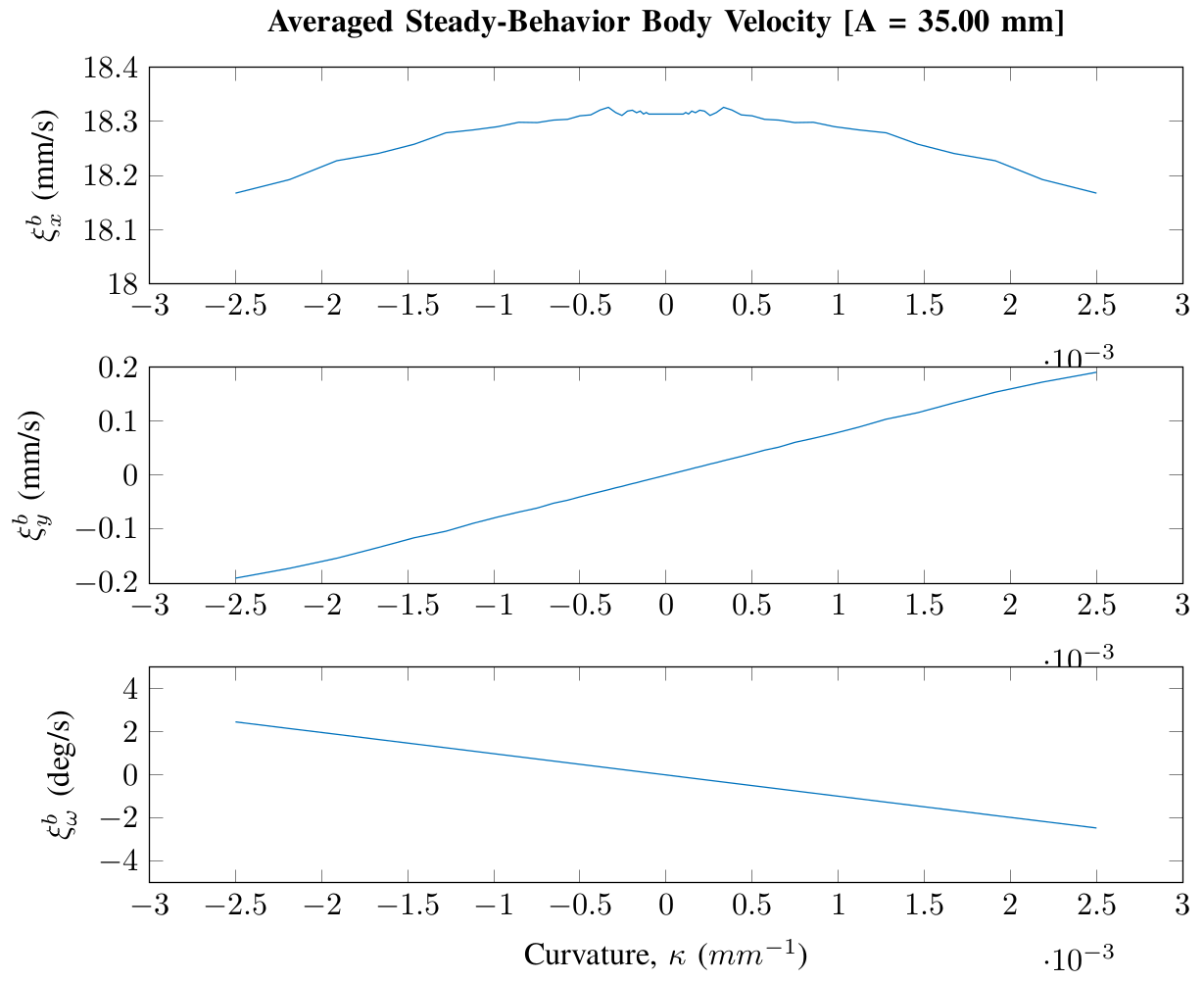}}
  \caption{\textbf{Body Velocity vs Curvature, $\body \xi(\kappa)$}:
    Linear components (\textbf{top} and \textbf{middle}) remain relatively
    constant with respect to average body curvature. 
    The angular component \textbf{(bottom)} varies linearly with curvature. 
    \label{rectilinear_xi_w_vs_curvature}}
    \vspace{-6mm}
\end{figure}

Figure \ref{rectilinear_xi_w_vs_curvature} illustrates the resulting
relationship, generated from repeated numerical integration simulations
sweeping a wide range of traveling wave amplitude, $A$, and average body
curvature values, $\kappa$. Averaged steady-behavior body velocity of the system
is captured as a function over gait parameter space, $\{A, \kappa\}$.
Linear body velocity components remained largely invariant with respect
to traveling wave amplitude, $A$, and average body curvature, $\kappa$.
Averaged steady-behavior body velocity laterally, $\xi^b_y$, is
negligible while that in the forward direction, $\xi^b_x$, remains
positive and nearly constant. 
Variation of traveling wave amplitude has little impact on the averaged
steady-behavior motion of the system.  Angular velocity varies
linearly with average body curvature, $\kappa$.

The linear relation between $\kappa$ and turn rate, as well as the
result that linear body velocity remains fairly constant, reveals a strong 
resemblance of this system to a fixed forward-velocity unicycle model, whereby
$\kappa$, serves as the steering control parameter. 
We denote the inverse mapping of $\body \xi_{\omega}(\kappa)$ by
$\kappa(\omega) = (\body \xi_{\omega})^{-1}(\omega)$ where $\omega$ is a
given angular velocity.  It is maintained as a linear fit of data plotted in Figure \ref{rectilinear_xi_w_vs_curvature}.
The mapping, $\body \xi(\kappa)$ effectively reduces the traveling wave
rectilinear gait dynamics to that of a kinematic unicycle vehicle, 
reminiscent of \cite{XuEtAl_ICRA[2015]}. Classical
techniques for path tracking are applicable;
body velocities required to rectify following errors are expeditiously mapped 
to corresponding body curvatures that produce the required motion.

\section{Perception and Planning}
\seclabel{planning}
A monocular radio frequency (RF) camera attached to the head of a
$12$-link robotic snake with scales \cite{SeEtAl_ICRA[2015]} supports
localization and planning.  
The camera continually transmits a $640\times480$ analog color image to
a stationary PC. All computation related to navigation, tracking and
control is processed on this PC with the final body shape commands
transmitted, through a lightweight tether, to the robot for locomotion.
The intent behind the wireless transmission method is to limit the
thickness of the tether to improve maneuverability.

The traveling wave rectilinear gait entails sinusoidal, time-varying
body shape changes over time. As a result, the camera frame is in
constant motion during the course of gait of execution, both spatially
and with respect to the robot body frame. Visual sensing of the
environment operates at one frame per gait cycle (i.e., every $2.5$ seconds),
coinciding with gait phase during which velocity of the camera frame is
minimal to reduce motion blur. 
Applying the unicycle motion model \cite{ChVe_ICRA[2017]} to the
previous pose estimate provides a predicted pose for the robot as a
function of the current commanded gait curvature ($\kappa$).
Each kept image frame and predicted pose is sent to a custom ORB-SLAM
implementation for estimating visual odometry during locomotion.  From
there, a vision-based path planner confirms the current trajectory or
returns a new path to follow for collision avoidance.


\subsection{Simultaneous Localization and Mapping (SLAM)}
%
%

ORB-SLAM \cite{MuTa_TRO[2017]} is chosen mostly due to the robustness of
ORB feature descriptor \cite{rublee2011orb}.  Other popular real-time
methods use direct visual odometry/SLAM
(e.g. SVO \cite{forster2017svo}, DSO \cite{engel2018direct}), which
relies on consistent illumination and low noise in the video input.  In
our platform, the assumptions required by direct methods are easily
violated due to the target application, the on-board sensor, and
transmission-based image corruption.
Feature-based ORB-SLAM is more robust to these nuisance factors:
it provides reliable odometry and 3D map points when working with rolling
shutter camera under image transmission induced noise.
One limitation of visual odometry/SLAM with monocular camera is that, 
it cannot provide accurate scale estimation since scale is not
observable given monocular video input.  
For visual odometry/SLAM to be useful in planning \& control, two
enhancements are added to monocular ORB-SLAM, enabling it to provide
well-bounded scale estimation.  

\begin{figure}[t!]
  \vspace{2.0mm}
  \centering
  \begin{tikzpicture}[inner sep=0pt, outer sep=0pt]
  \node (L) at (0in,0in)
    {\includegraphics[width=0.48\linewidth,bb=0 0 297 298]{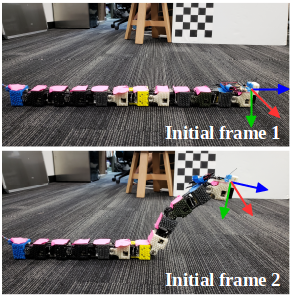}};
  \node[anchor=north west,xshift=2mm] (R) at (L.north east)
    {\includegraphics[width=0.44\linewidth,bb=0 0 480 524]{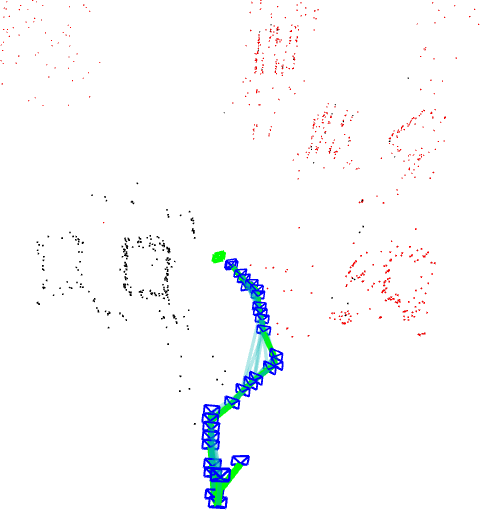}};

  \node (B) at ($(R.south east) + (-0.45in,0.175in)$) {};
  \draw (B) -- ($(B) + (-0.25in,0.025in)$);
  \draw (B) -- ($(B) + (-0.325in,-0.115in)$);
  \node[anchor=west] at (B) {\parbox{0.25in}{\footnotesize Initial poses}};

  \node (F) at ($(R.north west) + (0.25in,-0.3in)$) {};
  \draw (F) -- ($(F) + (0.40in,-0.45in)$);
  \node[anchor=south] at (F) {\parbox{0.25in}{\footnotesize Final pose}};

  \end{tikzpicture}
  \caption{\textbf{Left:} Initialization of ORB-SLAM with known-motion,
  wide-baseline frame capture.  \textbf{Right:} Key-frames and 3D map 
  generated with SLAM.  The scale of the estimated pose and map are accurate 
  enough for planning and control.
  \label{SLAM_InitExample}}
  \vspace{-11mm}
\end{figure}
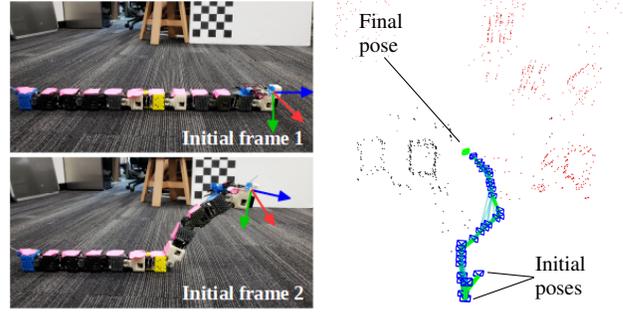

\boldpar{SLAM Initialization} Rather than employ the built-in, randomized
initialization strategy of ORB-SLAM, programmed snake movements with
known relative pose seeds the SLAM initialization.  With known head
movements between captured frames, see Fig.~\ref{SLAM_Results} left, the
unknown scale is fixed;
additionally, the initialization problem changes from a
structure-and-motion estimation to structure-only estimation for
initializing the map points.
The images used for SLAM initialization are captured with the widest baseline 
that the robot can provide through body movement (without tipping over),
thereby reducing the triangulation error of 3D mapped features, relative
to a typical motion-based initialization.  


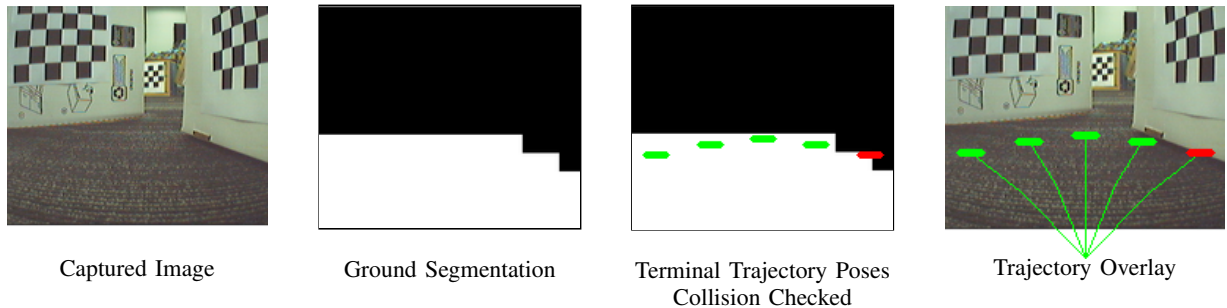
\begin{figure*}[t]
\vspace{2.5mm}
  \centering
  \begin{tikzpicture}[inner sep=0pt, outer sep=0pt]
    \node[anchor=north west] (F) at (0in,0in)
      {\includegraphics[width=0.4\columnwidth,bb=0 0 178 178,clip,trim=0in 0.25in 0in 0.125in]{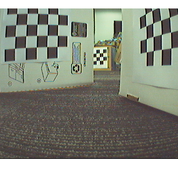}};
    \node[draw,very thick,inner sep=0pt,color=black,anchor=north west, xshift=2em] (B) at (F.north east)
      {\includegraphics[width=0.4\columnwidth,bb=0 0 178 178,clip,trim=0in 0.25in 0in 0.11in]{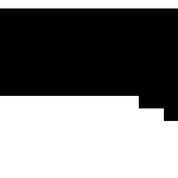}};
    \node[draw,very thick,inner sep=0pt,anchor=north west, xshift=2em] (P) at (B.north east)
      {\includegraphics[width=0.4\columnwidth,bb=0 0 178 178,clip,trim=0in 0.10in 0in 0.05in]{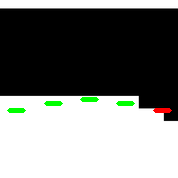}};
    \node[anchor=north west, xshift=2em] (O) at (P.north east)
      {{\includegraphics[width=0.43\columnwidth,bb=0 0 178 178,clip,trim=0in 0in
      0in 0.1in]{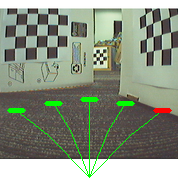}}};

    \node[anchor=north, yshift=-1.25em] at (F.south)
      {\small Captured Image};
    \node[anchor=north, yshift=-1.25em] at (B.south)
      {\small Ground Segmentation};
    \node[anchor=north, yshift=-1.25em] at (P.south)
      {\parbox{0.4\columnwidth}{\small \centering Terminal Trajectory
      Poses Collision Checked}};
    \node[anchor=north, yshift=-0.00em] at (O.south)
      {\small Trajectory Overlay};
  \end{tikzpicture}
  \caption{Perception space collision checking on binary
  ground/no-ground segmented image.  The captured image gets binarized,
  and poses representing the robot front are tested at discrete points
  along the trajectory.  Those leaving the binary region are considered
  collisions.  Trajectories whose poses stay full within the ground
  region are considered feasible. \figlabel{pips}}
  \vspace{-2.5mm}
\end{figure*}

\boldpar{Motion Model for Feature Matching Prior}
Though the scale is estimated accurately at the initialization, it
degenerates easily as the small estimation error accumulates in the
frame-by-frame tracking process of SLAM.  
We incorporate the snake unicycle motion model as the motion prior in
frame-by-frame tracking so as to bound the scale drift during the
tracking of ORB-SLAM.  
Only those features matchings that agree with the prior are accepted
and utilized in frame-by-frame tracking.  
Doing so reduces the estimation error and the scale drift during tracking
in ORB-SLAM.  The enhanced pose tracking and 3D mapping are illustrated in
Fig.~\ref{SLAM_InitExample} right.

\subsection{Perception Space Trajectory Planning}
Monocular, color vision does not recover dense depth of generic scenes,
while the monocular SLAM processing does not provide dense scene
geometry.  To overcome the lack of depth knowledge from monocular
streams, we employ ground-plane traversability tests to segment the
sensed environment into traversable and non-traversable components 
\cite{RaEtAl_TRO[2008],HoSe_TerrainTraversability,MaBe_2012_MonoTraversability,%
MeEtAl_OffRoadMonoTraversability,Goroshin2008_MonoObsDetect}.
When coupled with a flat ground-plane assumption, traversable regions can
be used to test the collision-free feasibility of potential forward
moving snake paths. For planning and trajectory synthesis to naturally
work with the segmented scene, it employs a perception space approach
\cite{SmVe_ICRA[2017]}.
Instead of relying on a 3D world reconstruction from the monocular
camera, the robot model is directly projected into image space for
collision checking.  The image itself is processed to obtain a binary
image where false values indicate hypothesized non-ground regions of the
world, e.g., obstacles, see Figure \figref{pips}.  The sample based
planner selects from amongst the collision-free paths to identify the
best path to follow.
Planning in binary perception space gives a flexible modality to deploy
our snake-like robotic platform for fast trajectory generation and
collision checking.

\boldpar{Ground Plane Segmentation}
Differentiating obstacles from ground uses a trained DCT-based support
vector machine \cite{HaVeCh_JCCE[2018]} to classify $40$-by-$40$ pixel
image blocks as \textit{ground} versus \textit{not ground}. The
prevailing assumption is that the camera pose will always have 
ground regions extending from the central bottom of the image up towards
the horizon, meeting with possible obstacle regions prior to hitting the
horizon.  Block-wise classification then yields a continuous
ground-obstacle boundary expanding from the bottom-center of the image.
A binary image is constructed whereby pixels below the boundary are
designated traversable and those above are impassible (ie. obstacles).
Figure \figref{pips} provides a visualization of the traversability
segmentation of an input image.


\boldpar{Collision Checking}  The planner uses a sample-based receding 
horizon strategy for synthesizing local paths and
testing their fitness.
In contrast to many path planning methods that assume a point based
robot model, our method considers the entire snake-like robot model
for performing collision checks.  Given a trajectory to collision check,
it is decomposed into several, closely-spaced navigation poses.
For each navigation pose, the robot model with $51$mm width front
is hallucinated, using the calibrated intrinsic camera matrix, and a
synthetic binary image is created for the robot head at that pose as
viewed from the current camera configuration.
Since there is a homographic assumption on the classified binary image
data, only the footprint of the snake robot head (no height geometry) is
hallucinated to create the future pose binary image.  Due to the
unicycle motion model, the snake body follows behind the head.
Collision checks at the head are also valid for the body.
%
The collision checking process involves evaluating the projected snake
head region and the binary image for false labels.
If the entire projected line fits in the ground plane area (only true
labels), the current pose will not collide with obstacles.  


\boldpar{Trajectory Sampling and Selection}
The trajectory sampler generates multiple trajectories to be collision
checked.  From the current pose of the robot, a set of $n$ curved
trajectories are generated based on near-identity diffeomorphism control
trajectories for unicycle robots \cite{SmVe_ICRA[2017]}, where $n$ is an odd 
number so as to include the straight trajectory (plus an equal quantity
going left as going right). We choose $n=5$ in our implementation. 
The chosen trajectory generation dynamics of the robot match the reduced
order kinematic model of the snake robot.
After checking all trajectory samples for collision, the longest
trajectory is chosen as the one to follow.


\section{Experiment}
\begin{figure}[t!]
\vspace{2mm}
  \centering
  {\includegraphics[width=1.0\columnwidth,bb=0 0 1600 1200,clip,trim=0in 0.25in 0in 0.125in]{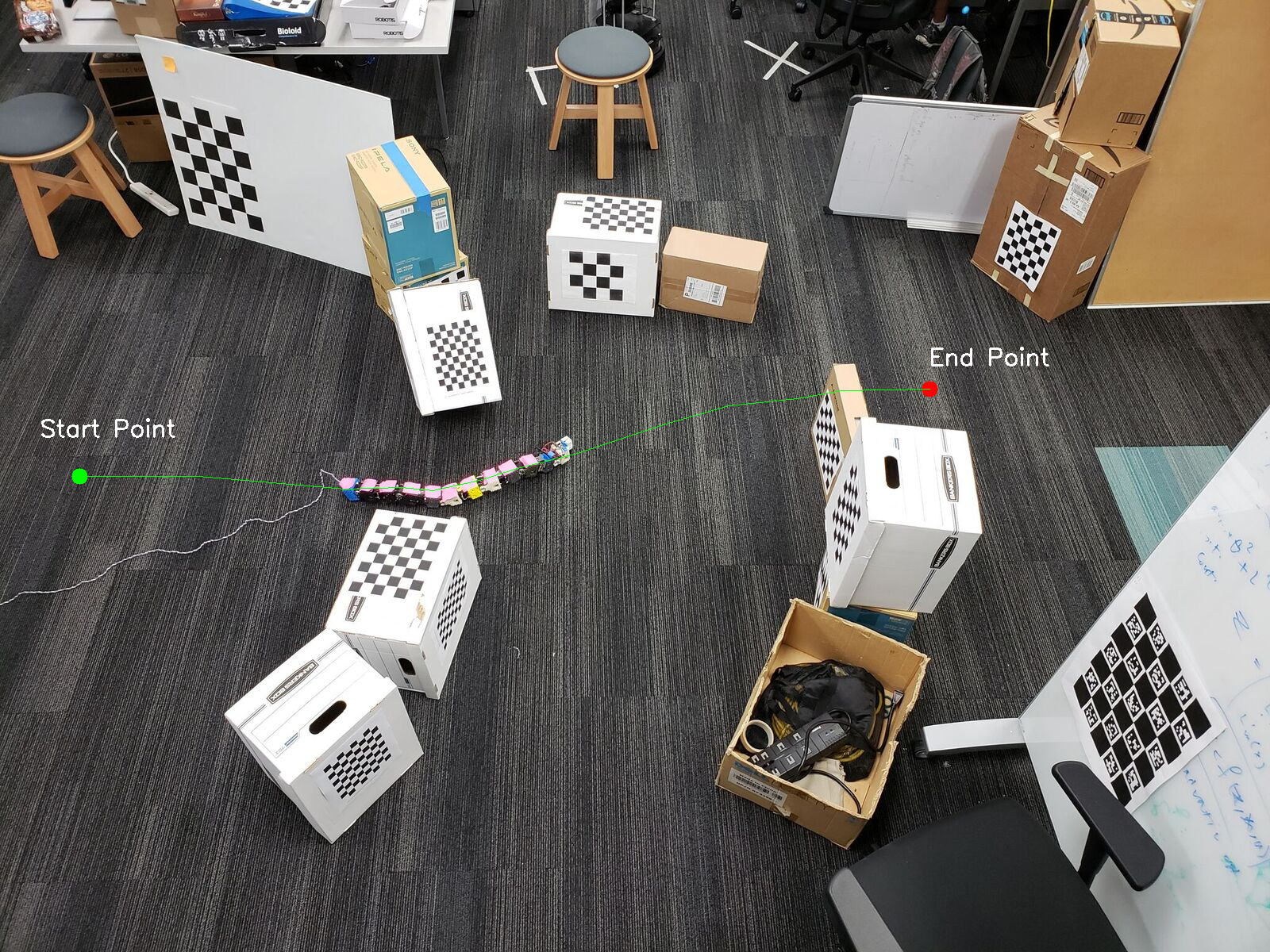}}
  \caption{Navigation overhead result of the snake-like robotic platform. The navigation
  system starts at the green point, and ends at the red point. Green curved line is the 
  real trajectory that robot walks through.
  \label{experim_scene}}
  \vspace{-3mm}
\end{figure}

\seclabel{results}
The presented navigation framework is experimentally deployed using a
$12$-link robotic snake operating over a carpeted surface.  We task the
robot to navigate through a non-trivial obstacle configuration,
illustrated in Figure \ref{experim_scene}.
Its initial objective is to travel a straight path along a corridor.
However, the obstacles in its path will induce a sequence of trajectory
re-plans as they are encountered.  The robot begins on the left and
attempts to proceed along a straight path from left to right. 

\begin{figure}[t!]
  \centering
  {\includegraphics[width=1.0\columnwidth,bb=0 0 1087 910]{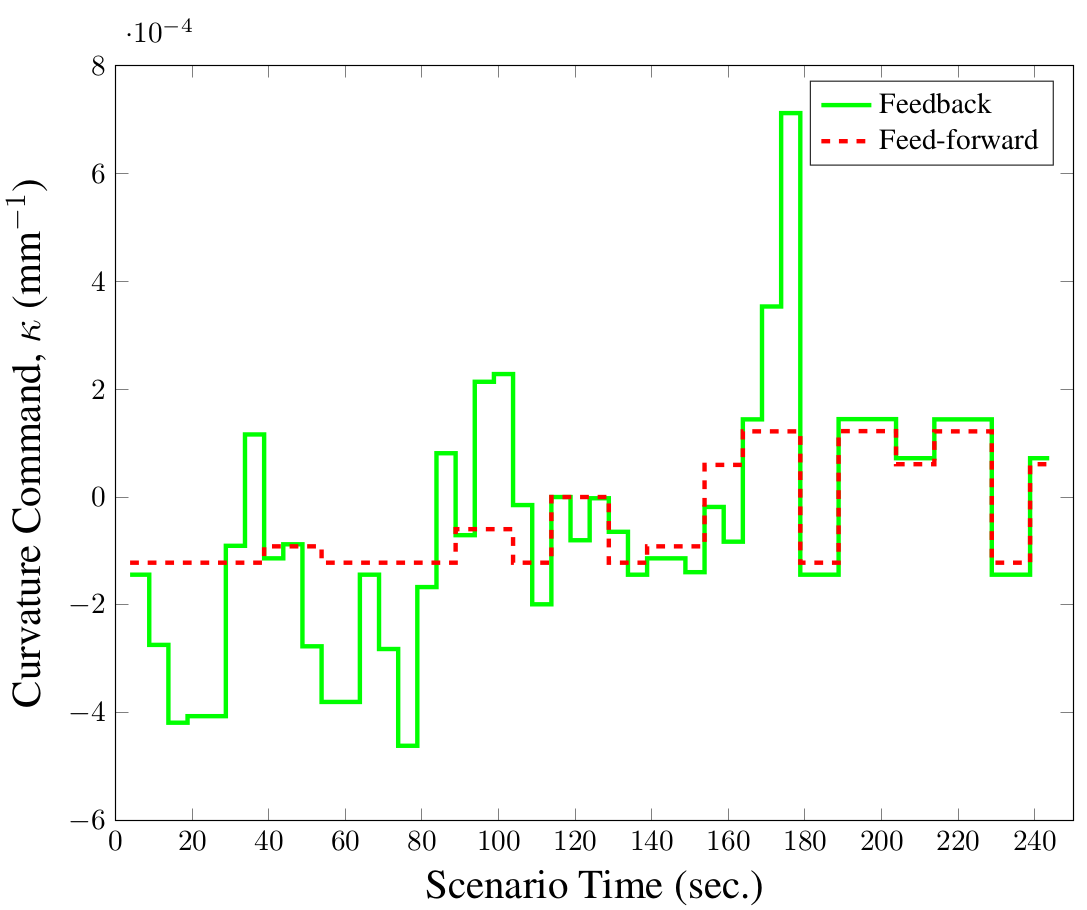}}  
  \caption{Body velocity corrections, to address robot pose errors with
  respect to the planned trajectory, are mapped back to feedback
  curvature commands (green) via the inverse control-to-action mapping,
  $\Phi$. Feed-forward curvature commands (red) of the planned trajectory
  are overlaid.
  \label{cntrl_curv_cmd}}
  \vspace{-3mm}
\end{figure}

As obstacles come into view, ground segmentation outputs cause the path
planning module to steer around them. A new path is re-planned every 
$10$ seconds (ie. every $4$ rectilinear gait cycles) based on the robot's
current position. Planned paths are represented as a sequence of
waypoints. The furthest waypoint, within a ball of radius  $\delta =
60$mm, is chosen as the target waypoint to which to travel. Robot pose
errors with respect to this waypoint are used to compute an angular body
velocity correction, $\omega_{fb}$ (as foward velocity is constant), to
the feedfoward angular velocity, $\omega_{ff}$, associated with the
planned trajectory. The control-to-action mapping, $\Phi$, then maps the
feedback corrected velocity to actionable curvature command, $\kappa$.
Figure \ref{cntrl_curv_cmd} illustrates the feedback-corrected curvature
commands versus the original feedforward cuvature commands computed
during the course of navigating the scenario of Figure \ref{experim_scene}. 
The feedback control strategy employed is similar to that used in
\cite{ChVe_ICRA[2017], ChEtAl_ACC[2018]}.

\begin{figure*}[t!]
\vspace{2mm}
  \centering
  \includegraphics[width=0.75\textwidth,bb=0 0 434 343]{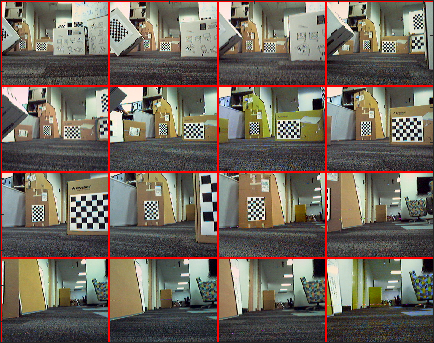}
  \caption{Visual information from the monocular head camera of the robotic snake informs both ORB-SLAM as well as path planning in the course of avoiding obstacles while traveling toward the goal. Images, captured during the course of the robot's trajectory through the obstacle field, is time-lapsed. The scenario begins in the top-left image; time proceeds from left-to-right, top-to-bottom.
  \label{egoSnapshots}}
  \vspace{-2mm}
\end{figure*}

\begin{figure}[t!]
  \centering
  \includegraphics[width=0.6\linewidth,bb=0 0 676 525]{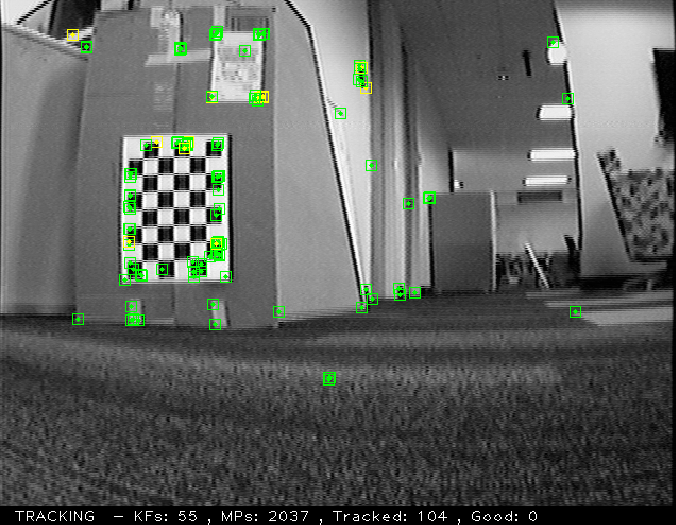}
  \includegraphics[width=0.38\linewidth,bb=0 0 390 513]{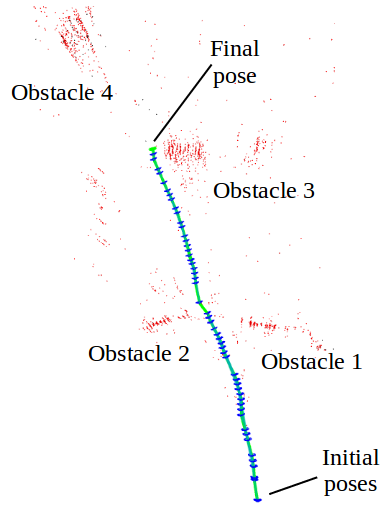}
  \caption{\textbf{Left:} Feature matching results of ORB-SLAM.  Inlier matchings 
  between current image and the local map are plotted with green boxes (104 in total); 
  while the outliers are in yellow boxes (10 in total).  
  \textbf{Right:} Top view of key-frames and 3D map generated with SLAM.  
  Notice how the trajectory of the robot goes around obstacles.
  \label{SLAM_Results}}
      \vspace{-2mm}
\end{figure}

A time-lapsed series of images, captured from the head camera, are
compiled in Figure \ref{egoSnapshots}. Despite a limited field of view
of the surrounding environment and control constraints on the turning
rate of the robot snake, the candidate trajectories generated circumvent
obstacles. When multiple feasible trajectories are possible, the longest
trajectory with least curvature is selected amongst the candidates.
Feedback control then tracks the planned trajectory utilizing robot pose
updates that are updated using ORB-SLAM until trajectory re-planning
occurs. At that point, the newly selected trajectory segment is
followed.  The scenario completes when the robotic snake exits the obstacle
field into the corridor on the right side of Figure \ref{experim_scene}.
From the visual sequence it is clear that the robot snake maneuvers to
navigate between the obstacles.  The camera pose of the robot tracked by
ORB-SLAM is illustrated in Fig~\ref{SLAM_Results}, where the left image
provides an example viewpoint and its set of tracked feature points,
while the right side depicts the estimated trajectory.


\section{Conclusion}
This work leverages a previously presented feedback control model for
the traveling wave rectilinear gait of a snake-like robot. 
Reduction of the complex gait dynamics to a simpler kinematic unicycle
model, in the steady-behavior motion, admits the application of tools and
strategies targeted for differential drive vehicles; in particular, to
trajectory planning and to trajectory tracking using feedback control.
We augment the snake-like robotic platform with a wireless monocular head
camera to capture images of the environment once every gait cycle.
Future pose predictions derived from the gait motion model, in
conjunction with captured visual information, inform ORB-SLAM which is tasked
with self-localization in the environment. Ground segmentation of
captured visual information then aids a perception space trajectory
planner.  Localization is critical to successful tracking of planned
trajectories. 
The integrated snake localization and navigation system is experimentally
deployed.  
The deployment self-localizes and dynamically plans through an unknown
environment as visual information of the environment becomes available.
Initially feasible trajectories are re-synthesized to navigate around
detected obstacles.  Using the presented navigation framework the robot
successfully negotiated a scenario, avoiding obstacles as they came into
view, and eventually exited the obstacle field.  Future work aims to
resolve the limitations of monocular cameras, both for localization and
navigation, by using a stereo camera and appropriately upgrading the
algorithmic system components.

\balance 

\bibliographystyle{IEEEtran}
\bibliography{regular,robosnake,biomimetic,ivalab,icra_list_refs,cvSLAM}

\end{document}